\documentclass{article}

\usepackage{arxiv}

\usepackage[utf8]{inputenc} % allow utf-8 input
\usepackage[T1]{fontenc}    % use 8-bit T1 fonts
\usepackage{hyperref}       % hyperlinks
\usepackage{url}            % simple URL typesetting
\usepackage{booktabs}       % professional-quality tables
\usepackage{amsmath}        % math environments
\usepackage{amsfonts}       % blackboard math symbols
\usepackage{nicefrac}       % compact symbols for 1/2, etc.
\usepackage{microtype}      % microtypography
\usepackage{graphicx}
\usepackage{multirow}
\graphicspath{ {./images/} }

\title{Greedy Is a Strong Default: Agents as Iterative Optimizers}

\author{
 Yitao Li \\
  \texttt{yitaoli416@gmail.com} \\
}

\begin{document}
\maketitle
\begin{abstract}
Classical optimization algorithms---hill climbing, simulated annealing, population-based methods---generate candidate solutions via random perturbations. We replace the random proposal generator with an LLM agent that reasons about evaluation diagnostics to propose informed candidates, and ask: \emph{does the classical optimization machinery still help when the proposer is no longer random?}

We evaluate on four tasks spanning discrete, mixed, and continuous search spaces (all replicated across 3 independent runs): rule-based classification on Breast Cancer (test accuracy 86.0\%~$\to$~96.5\%), mixed hyperparameter optimization for MobileNetV3-Small on STL-10 (84.5\%~$\to$~85.8\%, zero catastrophic failures vs.\ 60\% for random search), LoRA fine-tuning of Qwen2.5-0.5B on SST-2 (89.5\%~$\to$~92.7\%, matching Optuna TPE with 2$\times$ efficiency), and XGBoost on Adult Census (AUC 0.9297~$\to$~0.9317, tying CMA-ES with 3$\times$ fewer evaluations). Empirically, \textbf{on these tasks}: a cross-task ablation shows that simulated annealing, parallel investigators, and even a second LLM model (OpenAI Codex) provide no benefit over greedy hill climbing while requiring 2--3$\times$ more evaluations. In our setting, the LLM's learned prior appears strong enough that acceptance-rule sophistication has limited impact---round~1 alone delivers the majority of improvement, and variants converge to similar configurations across strategies. The practical implication is surprising simplicity: greedy hill climbing with early stopping is a strong default. Beyond accuracy, the framework produces human-interpretable artifacts---the discovered cancer classification rules independently recapitulate established cytopathology principles. Code is available at \url{https://github.com/yitao416/agentic-optimization}.
\end{abstract}

\keywords{LLM agents \and optimization \and hyperparameter tuning \and proposal distribution}

\section{Introduction}

Hyperparameter optimization and artifact optimization are fundamentally iterative search problems. Current approaches---random search, Bayesian optimization, evolutionary strategies (CMA-ES), grid search---rely on random or model-based perturbations to generate candidate solutions. A growing body of work uses large language models as optimizers: OPRO~\cite{yang2024opro} iterates on prompts, EvoPrompting~\cite{chen2023evoprompting} combines LLMs with evolutionary search, FunSearch~\cite{romeraparedes2024funsearch} pairs LLMs with evolutionary procedures, and LLAMBO~\cite{liu2024llambo} integrates LLMs into Bayesian optimization.

These systems share a common structure: \emph{propose a candidate, evaluate it, decide whether to accept}. This is precisely the structure of classical optimization algorithms---hill climbing, simulated annealing, and population-based methods. The only difference is the \textbf{proposal generator}: an LLM that reads evaluation diagnostics and proposes changes based on reasoning, rather than a random perturbation. This raises a natural question: \emph{when the proposal generator is an informed LLM rather than a random process, does the surrounding optimization machinery---simulated annealing, parallel search, strategic diversity---still provide value?}

We investigate this question empirically on four tasks spanning discrete rule discovery, mixed hyperparameter optimization, and continuous tuning. Across all tasks and 3 independent runs per configuration, we find that greedy hill climbing matches or beats simulated annealing and parallel investigators while using 2--3$\times$ fewer evaluations. The LLM's learned prior appears to be the dominant factor: round~1 alone delivers the majority of improvement, and all variants converge to similar configurations regardless of acceptance strategy.

We do not claim this holds universally---on tasks with sharper local optima or weaker LLM priors, SA may still help. But the consistency of the finding across four diverse tasks suggests that, at least for current LLMs on standard ML optimization problems, \textbf{the proposal quality matters far more than the acceptance rule}. The practical implication is that researchers building LLM-based optimizers can start with the simplest possible loop---greedy acceptance with early stopping---and add complexity only if needed. The framework also produces practically useful results: matching Optuna TPE and CMA-ES with fewer evaluations, zero catastrophic failures, and human-interpretable artifacts that independently recapitulate domain knowledge.

\section{Framework}
\label{sec:framework}

\subsection{Iterative Optimization as Propose-Evaluate-Accept}

Let $\theta$ denote the current artifact (a rule file, hyperparameter configuration, or any parameterized object) and $M(\theta)$ its evaluation metric. Each optimization round proceeds as:

\begin{enumerate}
\item \textbf{Propose:} Generate a candidate $\theta'$. Classically, $\theta'$ is drawn from a random perturbation distribution around $\theta$. In our framework, $\theta'$ is proposed by an LLM agent that reasons about evaluation diagnostics.
\item \textbf{Evaluate:} Compute $M(\theta')$ on the candidate.
\item \textbf{Accept/Reject:} Decide whether to adopt $\theta'$ based on an acceptance rule.
\end{enumerate}

The acceptance rule determines the optimization algorithm:

\subsection{Hill Climbing (Greedy Acceptance)}

Accept $\theta'$ if $M(\theta') > M(\theta)$; reject otherwise. This converges to a local optimum. In our implementation, the Orchestrator dispatches the Investigator to propose $\theta'$, and the Reviewer accepts if the metric strictly improves.

\subsection{Simulated Annealing (Probabilistic Acceptance)}

Accept improvements always; accept regressions with probability:
\[
P(\text{accept}) = \exp\!\left(\frac{M(\theta') - M(\theta)}{T}\right)
\]
where $T$ is the temperature, cooled geometrically: $T_k = T_0 \cdot \gamma^{k-1}$. At high temperature, the optimizer explores freely; as $T \to 0$, it converges to greedy. Given a sufficiently slow cooling schedule, simulated annealing converges to the global optimum~\cite{kirkpatrick1983sa}. The implementation is identical to hill climbing except that the Reviewer applies the Boltzmann acceptance criterion.

\subsection{Parallel Investigators}

$K$ investigators run independently each round, each receiving a \emph{directive} constraining it to a different search region (e.g., ``explore deep trees with heavy pruning'' vs.\ ``explore class-aware calibration''). The Reviewer selects the best proposal. This is parallel hill climbing with selection, where directives force proposal diversity beyond what the shared LLM prior provides.

\subsection{The LLM as Proposal Generator}

The Investigator reads diagnostic output---cross-validation results, error analysis, misclassification patterns---and proposes changes based on reasoning about observed patterns. For example, detecting overfitting leads to increased regularization; observing class imbalance triggers scale adjustment.

The key difference from random proposals is that the LLM has a \textbf{learned prior} over good configurations, derived from its training data. This prior is the source of both strength (efficient early search, informed categorical decisions) and weakness (convergence to ``conventional'' configurations, inability to match continuous optimizers on numerical precision).

Formally, let $\mathcal{L}$ denote the LLM (e.g., Claude Opus 4.6) and $f$ the deterministic parsing function that maps token sequences to valid artifacts. Given the current artifact $\theta$ and diagnostic information $d$, the model $\mathcal{L}$ produces a conditional distribution over token sequences; applying $f$ yields the proposal distribution over artifacts:
\[
P_{\mathcal{L}}(\text{tokens} \mid \text{prompt}(\theta, d)) \;\xrightarrow{\;f\;}\; q_{\mathcal{L}}(\theta' \mid \theta)
\]
The induced distribution $q_{\mathcal{L}}$ depends on the specific model $\mathcal{L}$---different LLMs will produce different proposal distributions and therefore different optimization trajectories. The effective support $\Theta_{q_{\mathcal{L}}} \subseteq \Theta$ (the region receiving non-negligible probability mass) is determined by $\mathcal{L}$'s training data and reflects its learned prior over good configurations. The LLM's value as a proposal generator is measured by whether $q_{\mathcal{L}}$ concentrates mass on improving solutions more effectively than a random perturbation distribution.

\subsection{When Does SA Help? A Proposal-Centric Analysis}
\label{sec:sa_theory}

Classical SA theory~\cite{kirkpatrick1983sa} shows that SA's advantage over greedy comes from crossing energy barriers between local optima: the Boltzmann criterion accepts uphill moves that enable the search to escape one basin and reach another. We argue that this advantage vanishes when the proposal distribution is sufficiently concentrated.

\paragraph{Effective search space.} For a random perturbation, $\Theta_q \approx \Theta$. For an LLM $\mathcal{L}$ with a strong prior, $\Theta_{q_{\mathcal{L}}} \ll \Theta$: proposals cluster in a narrow ``conventional'' region. Different models $\mathcal{L}_1, \mathcal{L}_2$ may have different effective supports $\Theta_{q_{\mathcal{L}_1}} \neq \Theta_{q_{\mathcal{L}_2}}$, but parallel investigators sharing the same $\mathcal{L}$ are constrained to the same $\Theta_{q_{\mathcal{L}}}$. We test this prediction empirically in Section~3.4 using two distinct LLMs.

\paragraph{Unimodality within $\Theta_{q_{\mathcal{L}}}$.} If the objective $M$ restricted to $\Theta_{q_{\mathcal{L}}}$ is unimodal---i.e., $\Theta_{q_{\mathcal{L}}}$ contains at most one local optimum---then greedy and SA converge to the same solution. SA's barrier-crossing mechanism requires multiple basins within the reachable set; if $\Theta_{q_{\mathcal{L}}}$ contains only one basin, there are no barriers to cross, and SA's stochastic acceptance adds noise without enabling escape.

\paragraph{Acceptance degeneracy.} Even when $\Theta_q$ contains mild multi-modality, a concentrated proposal distribution limits the magnitude of regressions. If $|M(\theta') - M(\theta)| \leq \varepsilon$ with high probability under $q$, then for any regression the SA acceptance probability satisfies:
\[
P(\text{accept}) = \exp\!\left(\frac{M(\theta') - M(\theta)}{T}\right) \geq \exp\!\left(\frac{-\varepsilon}{T}\right)
\]
When $\varepsilon / T$ is small (small regressions relative to temperature), $P(\text{accept}) \approx 1$ and SA accepts nearly all moves---degenerating to an undirected random walk over $\Theta_q$ rather than informed search. Greedy, by contrast, filters out regressions and converges monotonically.

\paragraph{Prediction.} This analysis predicts that SA underperforms or matches greedy when (1)~the LLM's effective support $\Theta_q$ is narrow, and (2)~$M$ is approximately unimodal within $\Theta_q$. Both conditions are testable from experimental data: condition~(1) corresponds to low proposal variance, and condition~(2) corresponds to all runs converging to similar final configurations. We evaluate these predictions in Section~\ref{sec:analysis}.

\subsection{Information Barriers}

The Investigator sees only training and cross-validation diagnostics---never test metrics. The Reviewer makes acceptance decisions based on validation or cross-validation scores. Test metrics are evaluated once at the end for reporting purposes only. This separation prevents overfitting to the test set and mirrors standard train/validation/test discipline in machine learning.

\section{Experiments}

We evaluate the framework on four tasks that differ in domain, artifact type, and search space structure. All experiments use Claude Opus 4.6 via the Claude Code CLI as the underlying LLM for the Investigator, Orchestrator, and Reviewer agents. Claude Code provides built-in tools for file editing, shell execution, and error investigation, as well as git worktree management that allows each Investigator to operate in an isolated copy of the repository---enabling parallel investigation without merge conflicts and clean rollback on rejection. Each Investigator is allowed up to 3 proposal attempts per round; if all attempts regress on the acceptance metric, the round is skipped and counts toward early stopping (2 consecutive skips trigger termination). All experiments enforce a strict 3-way evaluation protocol: the Investigator sees only training diagnostics, the Reviewer makes acceptance decisions on a held-out validation set (or cross-validation), and the test set is evaluated once at the end. This eliminates test-set leakage and ensures reported improvements reflect genuine generalization. Table~\ref{tab:summary} summarizes the results.

\begin{table}[t]
  \caption{Summary of results across all four experiments (best of 3 independent runs per task).}
  \centering
  \begin{tabular}{llllcc}
    \toprule
    Task & Artifact Type & Method & Metric & Seed & Best \\
    \midrule
    Cancer rules & Discrete (YAML) & Greedy ($n\!=\!3$) & Test accuracy & 0.860 & 0.965 \\
    STL-10 pipeline & Mixed (16 params) & Greedy ($n\!=\!3$) & Test accuracy & 0.845 & 0.858 \\
    LoRA SST-2 & Mixed (8 params) & Greedy ($n\!=\!3$) & Test accuracy & 0.895 & 0.927 \\
    XGBoost Adult & Continuous (12 params) & Full ablation & Test AUC & 0.9297 & 0.9317 \\
    \bottomrule
  \end{tabular}
  \label{tab:summary}
\end{table}

\subsection{Experiment 1: Discrete Rule Discovery---Breast Cancer}

\paragraph{Task.} Iteratively discover YAML classification rules for the Breast Cancer Wisconsin dataset (569 samples, 30 features, binary classification: malignant vs.\ benign). Rules consist of conditions on numeric features with thresholds and logical combinators, evaluated top-to-bottom with first-match semantics.

\paragraph{Setup.} A clean 3-way stratified split (341 train / 114 validation / 114 test, random\_state=42) was used. Seed rules (3 rules on worst radius and concave points) achieved 88.0\% train / 89.5\% validation / 86.0\% test. Six independent runs were conducted: 3 greedy, 3 simulated annealing (SA; $T_0 = 0.054$, $\gamma = 0.7$, calibrated so that a 1-sample validation regression has $p_{\text{accept}} \approx 0.85$ at round~1). Acceptance was based on validation accuracy; test was evaluated once at the end.

\paragraph{Results.} Table~\ref{tab:cancer} summarizes the replicated results. Greedy substantially outperforms SA: +1.8\% mean test accuracy, 3.6$\times$ lower standard deviation, and 2$\times$ fewer rounds.

\begin{table}[t]
  \caption{Breast Cancer rule optimization: 6 independent runs (3 greedy, 3 SA). Greedy achieves higher accuracy with lower variance and fewer rounds. SA's stochastic acceptance is counterproductive for this task.}
  \centering
  \begin{tabular}{lccccc}
    \toprule
    Method & Runs & Mean Test & Std & Best Test & Mean Rounds \\
    \midrule
    Seed & --- & 0.860 & --- & --- & 0 \\
    Greedy & 3 & \textbf{0.959} & 0.005 & \textbf{0.965} & 3.3 \\
    SA ($T_0\!=\!0.054$) & 3 & 0.941 & 0.018 & 0.956 & 6.7 \\
    \midrule
    sklearn DT (reference) & --- & 0.960 & --- & --- & --- \\
    sklearn RF (reference) & --- & 0.960 & --- & --- & --- \\
    \bottomrule
  \end{tabular}
  \label{tab:cancer}
\end{table}

The best greedy run (96.5\% test, 5 rules, 4 features) exceeds both sklearn Decision Tree and Random Forest (96.0\% each, trained on 25\% more data since they use train+val combined). A key structural insight emerged: in Greedy Run~2, the agent \emph{simplified} the rule set from 11 to 9 rules and 10 to 7 features; training accuracy dropped from 100.0\% to 98.2\%, but validation improved from 95.6\% to 97.4\%---the LLM performing implicit regularization. This simplification-as-improvement pattern recurred across runs.

SA's stochastic acceptance was counterproductive: across 3 SA runs, explore moves (accepting regressions) were mostly harmful. The first-round proposal quality dominates final performance (validation range 0.921--0.974 at round~1), and greedy correctly retains good proposals without the overhead of stochastic exploration.

\begin{figure}[t]
  \centering
  \includegraphics[width=0.65\columnwidth]{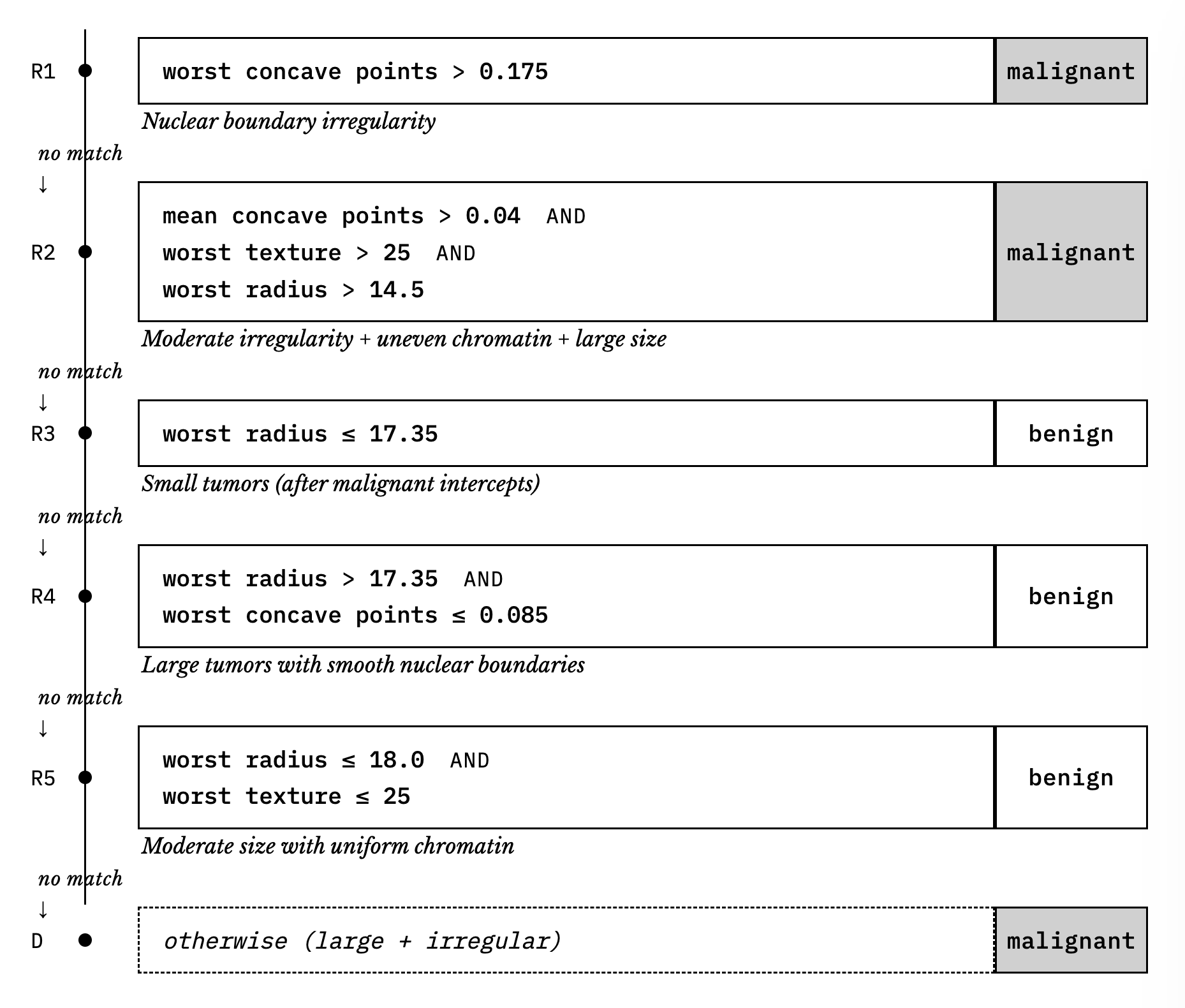}
  \caption{Agentic rule-based classifier for breast cancer diagnosis. Rules are evaluated sequentially (top to bottom); the first matching rule determines the classification. The system discovered 5 rules using 4 features from FNA biopsy measurements, achieving 95.6\% test accuracy with 341 training samples---comparable to a pruned decision tree (96\%) trained on 455 samples. Italic annotations are clinical interpretations based on the FNA cytopathology literature; they were not provided to the agent---the rules were discovered from data alone.}
  \label{fig:cancer_rules}
\end{figure}

Beyond accuracy, the framework produces \textbf{human-interpretable artifacts}. Figure~\ref{fig:cancer_rules} shows the simplest discovered model (Greedy Run~3: 5 rules, 4 features, 95.6\% test). Each rule maps to a recognizable clinical principle---the agent independently discovered that nuclear boundary irregularity (concave points) is the primary malignancy indicator, followed by chromatin texture and tumor size, mirroring established cytopathology. Unlike sklearn's 100-tree Random Forest or even its decision tree with numerically-optimized splits, the agentic rules are directly auditable by domain experts.

\subsection{Experiment 2: Mixed-Artifact Pipeline Optimization---STL-10}

\paragraph{Task.} Optimize 16 hyperparameters (optimizer type, learning rate schedule, data augmentation policy, training configuration) for a pretrained MobileNetV3-Small on STL-10 (4,000 train, 1,000 validation, 8,000 test images at 96$\times$96 resolution). The search space is mixed: categorical (optimizer type, augmentation policy), continuous (learning rate, weight decay), and conditional (optimizer-specific parameters).

\paragraph{Setup.} Three independent greedy runs from a common seed (SGD, lr=0.01, cosine schedule, crop+flip, no augmentation; validation 0.853, test 0.845). Acceptance was based on validation accuracy.

\paragraph{Baselines.} We compare against 30-trial random search and SMAC:

\begin{table}[t]
  \caption{STL-10: comparison against standard HPO baselines (30 trials each).}
  \centering
  \begin{tabular}{lccc}
    \toprule
    Method & Test Acc. & Trials & Failure Rate \\
    \midrule
    Default (seed) & 0.845 & 1 & 0\% \\
    SMAC & 0.840 & 30 & 33\% \\
    Random Search & 0.855 & 30 & 60\% \\
    \midrule
    Agentic (mean, $n\!=\!3$) & 0.852 & ${\sim}$8 & 0\% \\
    Agentic (best) & \textbf{0.858} & ${\sim}$4 & 0\% \\
    \bottomrule
  \end{tabular}
\end{table}

Random search and SMAC exhibit high catastrophic failure rates (test accuracy $< 0.15$): 60\% and 33\% of trials respectively, due to degenerate hyperparameter combinations. The agentic method produced \textbf{zero catastrophic failures} across ${\sim}$25 total evaluations.

\paragraph{Results.} Table~\ref{tab:stl10} shows the three independent runs. Runs~1 and 3 converged to \emph{identical} 16-parameter configurations (AdamW, lr=0.0015, wd=0.015, cosine schedule, cifar10 auto-augment, cutout~8, mixup~0.1, label\_smoothing~0.1, batch~64), providing strong evidence that the LLM's proposals are knowledge-driven rather than stochastic. Run~2 diverged (retained SGD with one\_cycle scheduler), achieving lower accuracy.

\begin{table}[t]
  \caption{STL-10: three independent greedy runs. Runs 1 and 3 converged to identical 16-parameter configurations; Run 2 diverged to a different optimizer family. The best agentic result (0.858) beats random search (0.855) with zero catastrophic failures.}
  \centering
  \begin{tabular}{lcccc}
    \toprule
    Run & Rounds & Accepted & Val Acc. & Test Acc. \\
    \midrule
    Run 1 & 5 & 3 & 0.862 & \textbf{0.858} \\
    Run 2 & 3 & 1 & 0.853 & 0.841 \\
    Run 3 & 3 & 1 & 0.862 & \textbf{0.858} \\
    \midrule
    Mean & --- & --- & 0.859 & 0.852 \\
    \bottomrule
  \end{tabular}
  \label{tab:stl10}
\end{table}

The highest-impact changes were categorical decisions---SGD~$\to$~AdamW and no augmentation~$\to$~cifar10 auto-augment---that random search and SMAC explore uniformly. The LLM's domain knowledge also acts as an implicit constraint: it avoids degenerate parameter combinations, producing an IQR of ${\sim}0.02$ across all proposals compared to 0.60 for random search. Run~3 reached the optimal configuration in ${\sim}$4 training evaluations, 7.5$\times$ more efficient than random search's 30.

\subsection{Experiment 3: LoRA Fine-Tuning Optimization---SST-2}

\paragraph{Task.} Optimize 8 hyperparameters for LoRA (Low-Rank Adaptation) fine-tuning of Qwen2.5-0.5B on SST-2 binary sentiment classification (5,000 train, 1,000 validation, 872 test). The search space spans LoRA architecture (rank, alpha, dropout), optimizer (learning rate, weight decay, warmup ratio, scheduler), and training (batch size). Key parameter interactions include the alpha/rank ratio, which acts as a multiplier on the effective learning rate, and regularization stacking between dropout and weight decay.

\paragraph{Setup.} Three independent greedy runs from a common seed. Acceptance was based on validation accuracy.

\paragraph{Baselines.} We compare against 20-trial random search and Optuna TPE (Tree-structured Parzen Estimator~\cite{bergstra2011tpe}):

\begin{table}[t]
  \caption{LoRA SST-2: comparison against standard HPO baselines (20 trials each for Random and TPE).}
  \centering
  \begin{tabular}{lccc}
    \toprule
    Method & Val Accuracy & Test Accuracy & Trials \\
    \midrule
    Seed config & 0.866 & 0.895 & 1 \\
    Random Search & 0.905 & 0.921 & 20 \\
    Optuna TPE & 0.911 & 0.922 & 20 \\
    \midrule
    Agentic (mean, $n\!=\!3$) & 0.910 & \textbf{0.924} & ${\sim}$9 \\
    Agentic (best) & 0.910 & \textbf{0.927} & ${\sim}$12 \\
    \bottomrule
  \end{tabular}
\end{table}

\paragraph{Results.} All 3 runs achieved mean test accuracy 0.924 ($\pm$0.003), matching or exceeding Optuna TPE (0.922) with approximately half the training evaluations (${\sim}$9 vs.\ 20). Table~\ref{tab:lora_runs} shows individual runs.

\begin{table}[t]
  \caption{LoRA SST-2: three independent greedy runs. All runs match or exceed Optuna TPE (0.922 test) with ${\sim}$half the compute. The two-phase optimization pattern is highly reproducible.}
  \centering
  \begin{tabular}{lcccc}
    \toprule
    Run & Accepted & Best Val & Test & Training Runs \\
    \midrule
    Run 1 & 2 & 0.910 & \textbf{0.927} & ${\sim}$12 \\
    Run 2 & 2 & 0.911 & 0.922 & ${\sim}$8 \\
    Run 3 & 2 & 0.909 & 0.923 & ${\sim}$8 \\
    \midrule
    Mean & 2.0 & 0.910 & 0.924 & ${\sim}$9 \\
    \bottomrule
  \end{tabular}
  \label{tab:lora_runs}
\end{table}

The optimization exhibited a highly reproducible two-phase pattern across all 3 runs:

\textit{Phase~1 (Round~1):} The Investigator diagnosed under-convergence from training logs and applied standard LoRA heuristics---higher learning rate, cosine schedule, increased rank. This produced a +3.9--4.5\% validation gain in a single proposal, contributing the majority of total improvement.

\textit{Phase~2 (Round~2):} Regularization refinement via a qualitative regime shift: all 3 runs independently converged to alpha/rank=8 (alpha=64) with heavy dropout (0.2--0.25) and weight decay (0.1--0.15) to compensate for the higher effective learning rate. This coordinated multi-parameter change yielded +0.3--0.5\% additional validation gain. Table~\ref{tab:lora} shows the configuration evolution.

\textit{Phase~3 (Rounds~3--4):} Plateau; all runs triggered early stopping after 2 consecutive rejections.

\begin{table}[t]
  \caption{LoRA SST-2: hyperparameter evolution showing the two-phase optimization pattern. All 3 runs independently converged to the same alpha/rank=8 regime that Optuna TPE also discovered.}
  \centering
  \begin{tabular}{lcccc}
    \toprule
    Parameter & Seed & Phase 1 & Phase 2 (Final) & Optuna TPE \\
    \midrule
    LoRA rank & 8 & 16 & 8 & 8 \\
    LoRA alpha & 16 & 32 & 64 & 64 \\
    Alpha/rank & 2 & 2 & 8 & 8 \\
    Learning rate & 2$\times 10^{-5}$ & $10^{-4}$ & 2--3$\times 10^{-4}$ & 3.3$\times 10^{-4}$ \\
    Dropout & 0.05 & 0.05 & 0.20--0.25 & 0.28 \\
    Weight decay & 0.01 & 0.01 & 0.10--0.15 & 0.18 \\
    Effective LR & 4$\times 10^{-5}$ & 2$\times 10^{-4}$ & 1.6--2.4$\times 10^{-3}$ & 2.6$\times 10^{-3}$ \\
    \bottomrule
  \end{tabular}
  \label{tab:lora}
\end{table}

The convergence of all 3 independent runs to the same alpha=64 regime---which Optuna TPE also discovered independently---provides strong evidence that the LLM's proposals encode genuine domain knowledge about LoRA scaling, not random exploration.

\subsection{Experiment 4: Continuous HPO with Ablation---XGBoost on Adult}

\paragraph{Task.} Tune 12 XGBoost hyperparameters on the Adult Census Income dataset (${\sim}$48,842 samples, 14 features, binary classification with ${\sim}$3:1 class imbalance). All parameters are continuous or integer-valued. The primary metric is test AUC-ROC.

\paragraph{Setup.} Clean evaluation protocol: 80/20 stratified train/test split; 5-fold stratified cross-validation on the training set for acceptance decisions; test set evaluated once at the end. Five agentic variants were tested with 3 independent runs each: greedy, SA ($T_0 = 5 \times 10^{-4}$, $\gamma = 0.7$, 10 rounds; $T_0$ calibrated so that a typical CV regression has $p_{\text{accept}} \approx 0.85$ at round~1), parallel ($K\!=\!3$ investigators with strategic directives), and two multi-model variants using a second LLM (OpenAI Codex): \emph{Codex Advisory} (Opus proposes while incorporating mandatory Codex suggestions) and \emph{Codex-Driven} (Codex is the sole decision-maker; Opus only executes and evaluates).

\paragraph{Baselines and ablation comparator.} We compare against 50-trial runs of standard HPO methods, and include OPRO-lite as an ablation comparator to agentic runs:

\begin{table}[t]
  \caption{XGBoost Adult: standard HPO baselines plus OPRO-lite (ablation comparator for agentic runs).}
  \centering
  \begin{tabular}{lcc}
    \toprule
    Method & Test AUC-ROC & Evals \\
    \midrule
    Default & 0.9292 & 1 \\
    Random Search & 0.9299 & 50 \\
    Optuna TPE & 0.9301 & 50 \\
    OPRO-lite ($n\!=\!3$) & 0.9303 & 11 \\
    Grid Search & 0.9307 & 50 \\
    Optuna CMA-ES & 0.9317 & 50 \\
    \bottomrule
  \end{tabular}
\end{table}

\paragraph{Full ablation.} Table~\ref{tab:xgboost} summarizes the replicated results across all 15 agentic runs (5 variants $\times$ 3 runs).

\begin{table}[t]
  \caption{XGBoost Adult: full ablation across acceptance rules, parallelism, and multi-model variants (3 runs each). Greedy (single model) achieves the best mean test AUC with the fewest evaluations.}
  \centering
  \begin{tabular}{lcccc}
    \toprule
    Variant & Mean Test & Best Test & CV Evals & vs.\ CMA-ES \\
    \midrule
    \multicolumn{5}{l}{\emph{Single-model (Opus 4.6)}} \\
    \quad Greedy ($n\!=\!3$) & \textbf{0.9314} & \textbf{0.9317} & ${\sim}$13 & $-0.0003$ \\
    \quad SA ($n\!=\!3$) & 0.9313 & 0.9316 & ${\sim}$28 & $-0.0004$ \\
    \quad Parallel ($K\!=\!3$, $n\!=\!3$) & 0.9313 & 0.9314 & ${\sim}$33 & $-0.0004$ \\
    \midrule
    \multicolumn{5}{l}{\emph{Multi-model (Opus 4.6 + Codex)}} \\
    \quad Codex Advisory ($n\!=\!3$) & 0.9312 & 0.9313 & ${\sim}$15 & $-0.0005$ \\
    \quad Codex-Driven ($n\!=\!3$) & 0.9306 & 0.9309 & ${\sim}$15 & $-0.0011$ \\
    \midrule
    \multicolumn{5}{l}{\emph{Ablation comparator}} \\
    \quad OPRO-lite ($n\!=\!3$) & 0.9303 & --- & 11 & $-0.0014$ \\
    \bottomrule
  \end{tabular}
  \label{tab:xgboost}
\end{table}

\paragraph{Greedy is a strong default.} Greedy achieves the highest mean test AUC (0.9314) with the fewest evaluations (${\sim}$13). Its best run ties CMA-ES at 0.9317 with ${\sim}$15 evaluations---3$\times$ fewer than CMA-ES's 50. SA and parallel achieve virtually identical mean test AUC (0.9313) but require 2$\times$ and 2.5$\times$ more evaluations, respectively.

\paragraph{SA provides no benefit under clean evaluation.} With test-blind, CV-based acceptance, the difference between greedy and SA is 0.0001---well within noise. SA's Boltzmann criterion rarely triggered: across 30 total SA rounds (3 runs $\times$ 10 rounds), stochastic acceptance of regressions was infrequent and mostly harmful. This contrasts with preliminary experiments where test-based acceptance gave SA an artificial advantage by providing more rounds to implicitly select on test performance.

\paragraph{Parallel investigation provides no benefit.} Despite strategic directives forcing different search strategies, parallel ($K\!=\!3$) achieves the same mean test AUC as SA while using ${\sim}$33 evaluations. The ``class-aware'' directive won 4 of 6 accepted rounds across runs; the ``penalty-based'' directive never won. This suggests the LLM's shared prior dominates directive-induced diversity: all investigators converge to similar ``conventional'' configurations regardless of their assigned strategy.

\paragraph{Round~1 dominates.} Across all 9 agentic runs, round~1 contributed the majority of improvement (+0.0013--0.0020 CV gain), with subsequent rounds adding diminishing increments. All variants converge to similar configuration regions: depth 5--7, colsample 0.35--0.5, scale\_pos\_weight 2.5--4.0. CMA-ES finds a genuinely unconventional configuration (depth~8, $\gamma = 6.18$, reg\_lambda $= 10^{-4}$) outside the LLM's prior, explaining its slight edge in mean performance.

\paragraph{Multi-model variants: does a different LLM help?} To test whether the convergence to ``conventional'' configurations is an artifact of using a single model, we introduce a second LLM (OpenAI Codex) in two configurations. In \emph{Codex Advisory}, the Opus Investigator consults Codex each round and must adopt at least one suggestion; in \emph{Codex-Driven}, Codex is the sole decision-maker and its suggestions are applied verbatim. Codex Advisory (0.9312) performs comparably to Greedy (0.9314), confirming that the ${\sim}$0.931 ceiling is a property of the search space, not a single-model bias. However, Codex-Driven (0.9306) underperforms: unmoderated Codex suggestions are frequently too aggressive (e.g., $\gamma = 4.5$, reg\_lambda $= 7.8$), causing over-regularization that Opus would have tempered. The three variants form a clean ablation of \textbf{moderation quality}: Greedy (self-moderated, 0.9314) $>$ Advisory (Opus moderates Codex, 0.9312) $>$ Driven (no moderation, 0.9306). Codex does contribute novel ideas---asymmetric column sampling (colsample\_bytree $= 0.45$, colsample\_bylevel $= 0.95$) broke a plateau in one Advisory run---but these gains are offset by Codex's tendency toward multi-parameter overhauls that obscure which changes help.

\paragraph{OPRO-lite ablation: isolating agentic orchestration.} To distinguish the value of agentic orchestration from simple history-conditioned LLM proposal, we implement an OPRO-lite comparator inspired by OPRO~\cite{yang2024opro}: a single prompt-only proposer sees the top-10 past (config, CV score) pairs and outputs a new config directly, with greedy acceptance and 10 rounds (11 total evaluations). OPRO-lite achieves 0.9303 mean test AUC ($n\!=\!3$). The agentic greedy method outperforms OPRO-lite by +0.0011 (0.9314 vs.\ 0.9303) with comparable budget (${\sim}$13 vs.\ 11 evaluations). The gap traces to a specific failure: all 3 OPRO-lite runs set \texttt{scale\_pos\_weight} between 1.05 and 1.15, effectively ignoring the dataset's ${\sim}$3:1 class imbalance, while the agentic Investigator consistently discovers it from CV fold analysis and sets the parameter to 3--4. This isolates the contribution of \textbf{diagnostic reasoning}: reading training logs, analyzing per-fold metrics, and reasoning about parameter interactions---capabilities that history-conditioned prompting alone cannot provide.

\section{Analysis}
\label{sec:analysis}

\begin{figure*}[t]
  \centering
  \includegraphics[width=\textwidth]{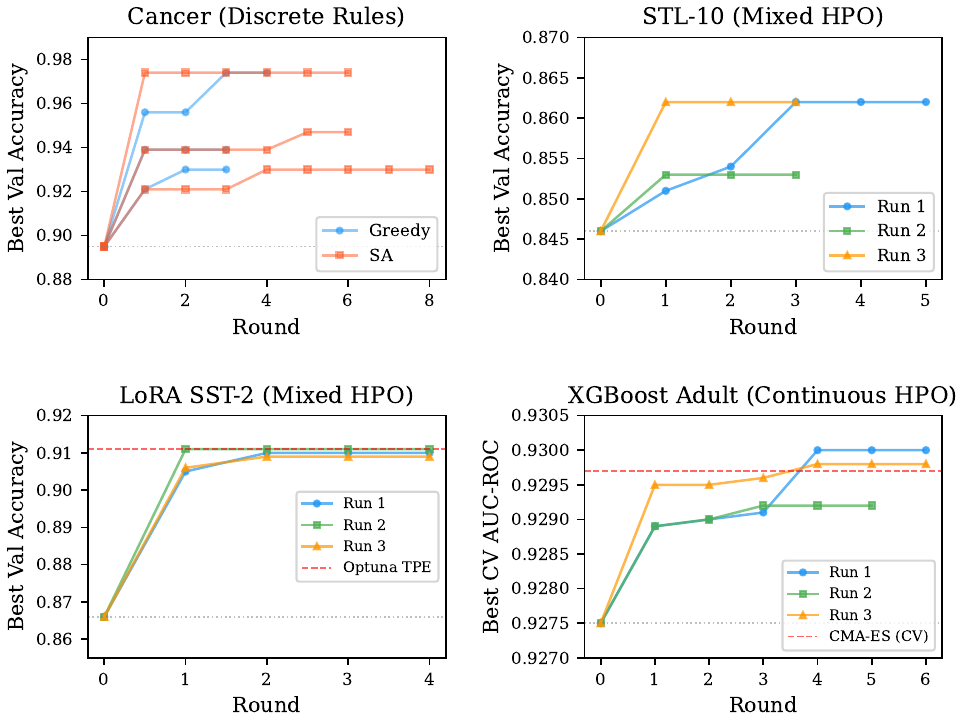}
  \caption{Best validation metric vs.\ optimization round across all four tasks (3 independent runs each). Round~1 delivers the majority of improvement in every case. Subsequent rounds add diminishing gains before early stopping. Cancer panel shows greedy (blue) vs.\ SA (orange); SA Run~2 achieved high validation but lower test accuracy (0.956 vs.\ greedy best 0.965), illustrating that validation-level advantage does not transfer to test.}
  \label{fig:convergence}
\end{figure*}

\subsection{When Does the LLM Proposal Generator Help?}

The LLM's advantage depends on the search space structure:

\paragraph{Discrete/combinatorial spaces (Cancer).} Massive advantage over random perturbation. Random rule modifications are nearly useless---they cannot reason about rule structure or feature interactions. The LLM understands the semantics of conditions and proposes meaningful modifications (e.g., tightening thresholds on boundary samples, adding conjunctive conditions, reordering rules to intercept malignant cases first). Result: Cancer 86.0\%~$\to$~96.5\% (best of 3 runs; mean 95.9\%), exceeding sklearn DT and RF (96.0\%) despite using 25\% less training data.

\paragraph{Mixed categorical-continuous spaces (STL-10, LoRA SST-2).} Clear advantage over random search and Bayesian methods. On STL-10, the LLM reasons about categorical choices (SGD~$\to$~AdamW, cifar10 auto-augment) and avoids catastrophic configurations, with 2 of 3 runs converging to identical solutions. On LoRA SST-2, the LLM reasons about compound parameter interactions (alpha/rank ratio, regularization stacking) to make a coordinated regime shift, matching Optuna TPE's result with 2$\times$ sample efficiency across all 3 runs.

\paragraph{Pure continuous spaces (XGBoost).} Competitive but not dominant. All agentic variants beat Random, TPE, and Grid with fewer evaluations. The best greedy run ties CMA-ES (0.9317 vs.\ 0.9317) with 3$\times$ fewer evaluations, but CMA-ES achieves a slightly higher mean by finding unconventional configurations outside the LLM's prior. In the OPRO-lite ablation (0.9303), the +0.0011 gap to agentic greedy (0.9314) shows that diagnostic reasoning---not just score history---is essential for discovering dataset-specific properties like class imbalance.

\subsection{The LLM Prior: Strength and Limitation}

The LLM's training data encodes optimization knowledge that acts as a strong prior over good configurations. This prior produces fast convergence: across all tasks, round~1 contributes the majority of total improvement (Figure~\ref{fig:convergence}). On LoRA SST-2, round~1 contributed +3.9--4.5\% validation gain per run, accounting for the majority of total improvement. On XGBoost, round~1 provided +0.0013--0.0020 CV gain, with subsequent rounds adding only diminishing increments.

However, the same prior limits late-stage optimization. All XGBoost investigators---whether greedy, SA, or parallel with directives---converge to the same ``conventional'' region: depth 5--7, moderate regularization, moderate class rebalancing. The CMA-ES solution (depth 8, $\gamma = 6.18$, $\lambda \approx 10^{-4}$) is genuinely unconventional and never discovered by any agentic variant.

This implies that \textbf{the LLM's proposal quality---not the acceptance rule---drives optimization performance}. The acceptance rule is a minor knob: switching from greedy to SA changes mean test AUC by 0.0001 on XGBoost, while the LLM's round~1 proposal contributes the majority of total improvement across all tasks. Directives provide structural diversity but cannot overcome the fundamental constraint that all proposals are drawn from the same learned prior. The multi-model experiment (Codex Advisory, 0.9312) confirms that the ${\sim}$0.931 ceiling is a search-space property: even a different LLM family cannot push past it. Meanwhile, the OPRO-lite ablation decomposes proposal quality: history-conditioned prompting achieves 0.9303, while agentic orchestration with diagnostic reasoning achieves 0.9314---a +0.0011 gap attributable to the Investigator's ability to analyze training logs and reason about parameter interactions rather than simply pattern-matching from score histories.

\subsection{Greedy as a Strong Default}

The cross-task ablation yields a clear recommendation for these settings ($n\!=\!3$ per condition): \textbf{greedy hill climbing with early stopping is a strong default} (Table~\ref{tab:cross_task}).

\begin{table}[t]
  \caption{Cross-task comparison of greedy vs.\ SA on the two tasks where both were tested. Greedy matches or beats SA with 2--3$\times$ fewer evaluations.}
  \centering
  \begin{tabular}{lccl}
    \toprule
    Task & Greedy Test & SA Test & SA Helps? \\
    \midrule
    Cancer (mean, $n\!=\!3$) & \textbf{0.959} & 0.941 & No ($-$0.018) \\
    XGBoost (mean, $n\!=\!3$) & \textbf{0.9314} & 0.9313 & No ($-$0.0001) \\
    \bottomrule
  \end{tabular}
  \label{tab:cross_task}
\end{table}

These results are consistent with the theoretical analysis in Section~\ref{sec:sa_theory}. Both predicted conditions hold: (1)~the LLM's effective support $\Theta_q$ is narrow---all XGBoost variants converge to depth 5--7, colsample 0.35--0.5 regardless of acceptance rule; and (2)~$M$ is approximately unimodal within $\Theta_q$---no run discovers a qualitatively different basin. With a concentrated proposal distribution and a unimodal effective landscape, SA's barrier-crossing mechanism has nothing to contribute. On tasks with sharper optima or weaker LLM priors, SA may still provide value.

\subsection{Reliability: Zero Catastrophic Failures}

\begin{figure}[t]
  \centering
  \includegraphics[width=\columnwidth]{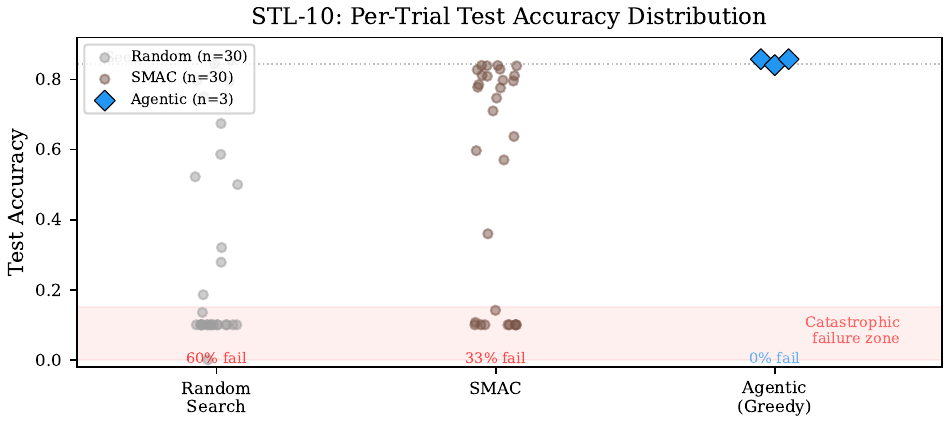}
  \caption{Per-trial test accuracy on STL-10 for random search, SMAC, and agentic greedy. Random search and SMAC produce catastrophic failures (test accuracy $< 0.15$) in 60\% and 33\% of trials. The agentic method produces zero failures across all evaluations, with tight clustering near the optimum.}
  \label{fig:reliability}
\end{figure}

Across all HPO experiments (${\sim}$25 STL-10 evaluations, ${\sim}$9 LoRA evaluations per run, ${\sim}$13--33 XGBoost evaluations per variant), the agentic method produced zero catastrophic failures (Figure~\ref{fig:reliability}) (test accuracy below chance level), compared to 60\% for random search and 33\% for SMAC on STL-10. The LLM's domain knowledge implicitly constrains the search to viable regions: it pairs AdamW with appropriate learning rates, avoids incompatible scheduler-optimizer combinations, and adjusts regularization parameters coherently rather than independently.

\subsection{Efficiency Comparison}

Table~\ref{tab:efficiency} compares evaluation budgets across methods.

\begin{table}[t]
  \caption{Evaluation efficiency: agentic greedy achieves competitive or superior results with fewer evaluations. OPRO-lite is shown as an ablation comparator for XGBoost and falls short of the agentic method.}
  \centering
  \begin{tabular}{lcccc}
    \toprule
    Task & Agentic Evals & Baseline Evals & Agentic Result & Best Baseline \\
    \midrule
    Cancer & ${\sim}$3 & N/A & 0.965 & 0.960 (DT/RF) \\
    STL-10 & ${\sim}$4--12 & 30 & 0.858 & 0.855 (random) \\
    LoRA SST-2 & ${\sim}$9 & 20 & 0.927 & 0.922 (TPE) \\
    XGBoost (greedy) & ${\sim}$13 & 50 & 0.9317 & 0.9317 (CMA-ES) \\
    XGBoost (OPRO-lite ablation) & 11 & --- & 0.9303 & 0.9317 (agentic greedy) \\
    \bottomrule
  \end{tabular}
  \label{tab:efficiency}
\end{table}

\section{Related Work}

\paragraph{Bayesian optimization.} TPE~\cite{bergstra2011tpe} and Gaussian process-based methods~\cite{snoek2012practical} model the objective function to guide search. These are complementary to our approach: the LLM could serve as the acquisition function's proposal generator rather than replacing the entire pipeline.

\paragraph{Evolutionary strategies.} CMA-ES~\cite{hansen2001cmaes} adapts a multivariate Gaussian proposal distribution using the covariance structure of successful candidates. Our XGBoost results show that CMA-ES's learned covariance can discover unconventional configurations outside the LLM's prior, achieving a slight edge in mean performance on continuous spaces.

\paragraph{LLM-based optimization.} OPRO~\cite{yang2024opro} uses LLMs to iteratively optimize prompts by including past solutions in the context. EvoPrompting~\cite{chen2023evoprompting} combines LLM-generated code with evolutionary algorithms. FunSearch~\cite{romeraparedes2024funsearch} pairs an LLM with an evolutionary procedure to discover new mathematical constructions, searching in program space rather than solution space. Our framework generalizes these: OPRO is a special case with full-history context and no acceptance gate; EvoPrompting and FunSearch correspond to population-based methods with LLM-generated mutations. Unlike FunSearch, which maintains a population of programs, our framework uses a single current-best artifact with classical acceptance rules. We include an OPRO-lite baseline (Section~3.4) to isolate the contribution of agentic orchestration: the agentic method outperforms history-conditioned prompting by +0.0011 AUC on XGBoost, attributable to diagnostic reasoning about training logs and per-fold metrics.

\paragraph{LLM-enhanced Bayesian optimization.} LLAMBO~\cite{liu2024llambo} integrates LLMs into Bayesian optimization by framing surrogate modeling and candidate sampling in natural language. LLAMBO improves BO performance especially in early search when observations are sparse, and is modular enough to enhance existing BO frameworks. Our approach is complementary but structurally different: we replace the entire proposal mechanism with an LLM agent rather than enhancing individual BO components, and we use greedy acceptance rather than acquisition functions.

\paragraph{LLM-assisted AutoML.} Recent systems use LLMs for automated machine learning and feature engineering~\cite{hollmann2023caafe}, neural architecture search, and pipeline optimization. Our contribution is positioning the LLM specifically as the proposal generator within classical optimization theory, rather than building a new end-to-end system.

\paragraph{Agentic coding tools.} Our framework is implemented using Claude Code~\cite{anthropic2025claudecode}, an agentic coding tool that provides the LLM with file editing, shell execution, and git worktree management capabilities. The Investigator agents operate in isolated git worktrees, enabling parallel exploration without merge conflicts and clean rollback on rejection. This tool infrastructure is essential to the framework: the LLM's optimization value comes not just from reasoning about configurations, but from its ability to directly execute evaluations, inspect error logs, and iteratively refine artifacts in a real development environment.

\paragraph{Parameter-efficient fine-tuning.} LoRA~\cite{hu2022lora} enables efficient adaptation of large language models. Our LoRA SST-2 experiment demonstrates that the agentic framework extends naturally to LLM fine-tuning hyperparameter optimization, where the LLM proposal generator has strong prior knowledge about parameter interactions (e.g., alpha/rank scaling, regularization stacking).

\section{Conclusion}

We presented a framework that positions the LLM as a drop-in replacement for the proposal generator in classical optimization algorithms. The acceptance rule, cooling schedule, and convergence properties are inherited directly from the classical algorithm---the LLM contributes only informed proposals.

Across four experiments with replicated independent runs, the framework is most effective for discrete and mixed search spaces where the LLM's structural understanding provides a clear advantage over random perturbation (Cancer: 86.0\%~$\to$~96.5\% best, 95.9\% mean; STL-10: beats random and SMAC with zero catastrophic failures; LoRA SST-2: matches Optuna TPE with 2$\times$ efficiency). On pure continuous optimization, greedy achieves 0.9317 test AUC on the best XGBoost run, tying CMA-ES with 3$\times$ fewer evaluations.

The central finding from the full ablation on XGBoost---spanning acceptance rules (greedy, SA), parallelism (directives), multi-model proposals (Codex Advisory/Driven), and an OPRO-lite comparator---is that \textbf{greedy hill climbing is a strong default in our tested settings}. SA, parallel investigators, and even a second LLM model provide no meaningful benefit while requiring more evaluations. The multi-model experiment confirms the performance ceiling is a search-space property, not a single-model artifact. The OPRO-lite ablation (+0.0011 gap to agentic greedy) isolates the value of diagnostic reasoning: the ability to read training logs and analyze per-fold metrics, rather than simply conditioning on score histories.

This leads to a clean conceptual picture: \textbf{the LLM's proposal quality dominates the choice of acceptance rule}. The iterative loop enables multi-round refinement, but the LLM's prior over good configurations does the heavy lifting. Classical acceptance rules (SA, parallel tempering) assume the proposal distribution is uninformed and needs correction; when the proposer is an LLM with strong domain knowledge, the correction adds overhead without benefit.

The practical implication is simplicity: a greedy hill climber with 2--3 rounds and early stopping captures the LLM's value. The framework's strengths---structured search, categorical reasoning, domain knowledge, zero catastrophic failures---emerge from the LLM's learned prior, with the iterative evaluation loop providing the feedback signal for refinement.

\paragraph{Limitations.} Several limitations merit discussion.
\emph{Model diversity.} The multi-model experiment (Codex Advisory/Driven) tests heterogeneous LLMs on XGBoost but finds no performance gain---the ceiling appears to be a search-space property. However, this was tested on only one task; on problems where the single-model prior is more limiting (e.g., novel architectures with no web-scale training data), model diversity may be more valuable.
\emph{Cost.} We compare evaluation counts but not total cost: each round incurs LLM API calls in addition to training evaluations, which may exceed the cost of additional baseline evaluations for cheap-to-evaluate objectives.
\emph{Statistical power.} With $n\!=\!3$ runs per condition, differences smaller than ${\sim}$0.005 are not statistically distinguishable.
\emph{Scale.} All search spaces are modest (8--16 parameters); scalability to higher-dimensional spaces is untested.
\emph{Memorization.} The LLM's convergence to ``conventional'' configurations raises the question of whether its advantage reflects genuine optimization reasoning or memorization of known-good configurations from training data---a distinction we cannot resolve with the current experimental design.

\bibliographystyle{unsrt}

\end{document}